\documentclass{article}

\usepackage{arxiv_preprint,times}

\usepackage{amsmath,amsfonts,bm}

\def\eqref#1{equation~\ref{#1}}

\def\1{\bm{1}}

\DeclareMathAlphabet{\mathsfit}{\encodingdefault}{\sfdefault}{m}{sl}
\SetMathAlphabet{\mathsfit}{bold}{\encodingdefault}{\sfdefault}{bx}{n}

\usepackage{amsmath}
\usepackage{amssymb}
\usepackage{booktabs}
\usepackage{tabularx}
\usepackage{graphicx}
\usepackage{placeins}
\usepackage{microtype}
\usepackage{etoolbox}
\usepackage{url}

\usepackage{hyperref}

\AtBeginEnvironment{table}{%
  \setlength{\abovecaptionskip}{0pt}%
  \setlength{\belowcaptionskip}{10pt}%
}
\AtBeginEnvironment{table*}{%
  \setlength{\abovecaptionskip}{0pt}%
  \setlength{\belowcaptionskip}{10pt}%
}

\newcommand{\runinlabel}[1]{\noindent\textbf{#1}\ }

\title{The Right Fact, the Wrong Decision:\\ SciValidityBench for Scientific Agent Memory}

\usepackage{fontawesome5}

\author{%
  \textbf{Yuexi Pan}$^{1,2}$ \qquad \textbf{Yongchao Chen}$^{1}$\thanks{Corresponding author.}\\
  \normalfont $^{1}$Tsinghua University\\
  \normalfont $^{2}$University of Science and Technology of China\\
  \normalfont\texttt{panyuexi@mail.ustc.edu.cn}\quad
  \texttt{chenyongchao@mail.tsinghua.edu.cn}\\[5pt]
  \normalfont\small\faGithub\quad
  \href{https://github.com/SciValidityBench/SciValidityBench}{GitHub: SciValidityBench/SciValidityBench}\\[3pt]
  \normalfont\small\faGlobe\quad
  \href{https://scivaliditybench.github.io/SciValidityBench/}{Website: scivaliditybench.github.io/SciValidityBench}%
}

\makeatletter
\renewcommand{\@maketitle}{%
  \vbox{%
    \hsize\textwidth
    \centering
    {\normalfont\LARGE\bfseries\@title\par}%
    \vskip 12pt
    {\normalfont\normalsize
      \begin{tabular}[t]{c}\@author\end{tabular}\par}%
    \vskip 0.1in minus 0.03in
  }%
}
\makeatother
\renewcommand{\headrulewidth}{0pt}
\hypersetup{%
  pdftitle={The Right Fact, the Wrong Decision: SciValidityBench for Scientific Agent Memory},
  pdfauthor={Yuexi Pan and Yongchao Chen},
  pdfsubject={Scientific agent memory; arXiv preprint}%
}

\iclrfinalcopy
\begin{document}

\flushbottom

\maketitle

\begin{abstract}
Most agent-memory benchmarks frame long-term memory as retrieval: can an agent recover the relevant record from a long and evolving history? Scientific agents face a stricter problem: a correctly remembered fact may still be invalid to use. The same reported test result can support confirmation when the outcome and stopping rule were fixed in advance, yet only exploration when the outcome was selected from twenty inspected results. We call this \emph{scientific use validity}: whether remembered evidence is eligible to influence a decision under the state governing its interpretation. We introduce \textsc{SciValidityBench}, an executable benchmark factory with 64 TaskSpecs spanning five validity modules and six scientific domains. TaskSpecs couple generated experimental histories with scientific-state definitions, programmatic decision oracles, controlled interventions, and automatic scoring. The \textsc{SciStateForge} compiler--reader framework turns distributed histories into explicit decision-governing state. Counterfactual pairs and targeted corruption--repair controls then test whether models use those dependencies and localize failures. Across four model families, compiled scientific state improves valid decision-making by 9.2--18.8 percentage points over a generic Event Graph. Retrieval alone does not reproduce this cross-family gain, while content-identical flat and typed views show that the improvement comes from exposing governing state rather than a particular graph serialization. SciValidityBench thus shifts memory evaluation from finding the right fact to deciding whether that fact should control the decision.
\end{abstract}

\section{Introduction}
\label{sec:introduction}

An agent can recover the right fact from a laboratory history and still make
the wrong scientific decision. Existing long-term-memory benchmarks probe long
conversations, incremental interactions, state updates, selective forgetting,
and on-policy personalization
\citep{maharana2024locomo,wu2025longmemeval,hu2026memoryagentbench,
cheng2026amemgym}. Yet their central question remains whether an agent can
preserve, retrieve, or revise the relevant information. Scientific agents face
a second question: is that information eligible to govern the current decision?

Consider two notebooks that both truthfully report $p=0.03$. In one, the
researcher preregistered the outcome, fixed the sample size, and followed the
stopping rule. In the other, the researcher inspected twenty outcomes and
stopped after one crossed the significance threshold. The number is identical,
but its evidential status differs: the former may support a confirmatory claim,
whereas the latter remains exploratory. Preregistration fixes the analysis plan
before observation, while selective-inference methods account for data-dependent
selection \citep{nosek2018preregistration,fithian2014optimal}. Better retrieval
cannot resolve this distinction because the governing information is distributed
across the analysis plan, outcome family, access history, and stopping rule.
Returning the correct value without these dependencies can make an agent more
confidently wrong.

\begin{figure*}[t]
\centering
\includegraphics[width=\textwidth]{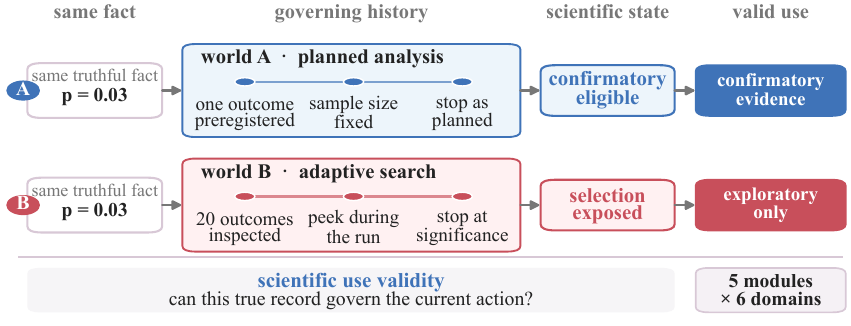}
\caption{\textbf{The right fact can license different decisions.} Retrieval
alone cannot distinguish the two worlds because the reported value is
identical. \textsc{SciValidityBench} evaluates whether an agent reconstructs
the cross-record state that determines how the evidence may be used. This
is the distinction between remembering a result and using it validly.}
\label{fig:right-fact-wrong-decision}
\end{figure*}

We call this missing evaluation object \emph{scientific use validity}
(Figure~\ref{fig:right-fact-wrong-decision}): whether remembered evidence may
influence an action under the scientific state governing its interpretation.
This problem also arises when an outcome
is viewed before a plan is locked, identically named measurements use different
constructs, a blank denotes failure rather than zero, or sunk costs influence a
prospective choice. Each case links a locally correct record to a decision
through state elsewhere in the history. Scientific memory must therefore
reconstruct the governing state, not merely retain the record.

This question is complementary to recent benchmarks for scientific agents.
Scientific-agent benchmarks test broad capabilities, controlled tools, and
open-ended or executable research workflows
\citep{bragg2026astabench,panigrahi2026heurekabench,kon2026expbench}, while
process evaluations show that final-answer accuracy can conceal unsupported
reasoning \citep{shao2026seekbench}. They do not isolate whether a memory reader
uses the state that licenses a particular decision. Structured memory may score
higher because of retrieval, added information, formatting, prompt length, or
answer preference. One correct answer alone cannot distinguish these causes.

We introduce \textsc{SciValidityBench}, an executable benchmark factory built
to make this distinction measurable. Each TaskSpec binds a generated
history, scientific-state definition, legal actions, programmatic oracle,
controlled interventions, and automatic scorer. The resulting 64 tasks cover
five validity mechanisms across machine learning, A/B testing, clinical
research, bioinformatics, systems, and causal inference. Domains change the
scientific surface and event generator, while mechanisms declare governing
state, interventions, and oracles through the same interface.
At the center of the benchmark is \textsc{SciStateForge}, a compiler--reader
pipeline that turns distributed experimental histories into explicit,
decision-governing scientific state. We compare complete raw events, fixed
BM25 retrieval, a generic Event Graph, a flat compiled-state view, and a typed
scientific-state ledger. Counterfactual worlds hold all non-target information
fixed while changing one decisive state value. Targeted corruption removes
oracle-required state. Exact and edit-matched irrelevant repairs distinguish
genuine state use from answer preference, generic text perturbation, and
irrelevant-history interference.

Across four locked model families, compiled scientific state improves
TaskSpec-balanced valid decision-making by 9.2--18.8 percentage points over a
generic Event Graph. BM25 retrieval does not reproduce this cross-family gain,
and content-identical flat and typed compiled views show that the central
benefit comes from exposing governing state rather than from a particular graph
serialization. Exact target repair exceeds matched irrelevant repair by 18.6
validity points on average. The resulting behavior identifies missing state
use, representation-sensitive shortcuts, and irrelevant-memory interference
rather than collapsing them into a single accuracy score. Our contributions
are threefold:

\noindent
\textbf{1. Scientific use validity.} We formalize the gap between recovering a
correct record and deciding whether it may govern an action, then instantiate
that distinction across five validity mechanisms with explicit state
dependencies.
\begingroup
\setlength{\parskip}{0pt}
\par\noindent
\textbf{2. An executable benchmark factory.} Sixty-four TaskSpecs across six
domains share generators, state contracts, legal actions, programmatic oracles,
interventions, and scorers, allowing new scientific settings to reuse the same
evaluation interface.
\par\noindent
\textbf{3. Causal diagnosis and evidence.} Paired worlds, targeted corruption,
exact repair, and edit-matched irrelevant repair separate genuine state use
from retrieval and answer preference. Compiled state beats Event Graph by
9.2--18.8 points across four model families.
\par
\endgroup

\section{Related Work}
\label{sec:related-work}

\runinlabel{Agent memory: from recalling facts to using them in action.}
Persistent-memory agents consolidate and selectively retain experience in
external stores \citep{park2023generative,packer2023memgpt}. Mem0 extends this
pattern to scalable deployments \citep{chhikara2025mem0}, and surveys map the
design space \citep{zhang2025memorysurvey}. Benchmarks cover long-range and
cross-session recall \citep{maharana2024locomo,wu2025longmemeval,tan2025membench,
tavakoli2026beam}, continual adaptation and forgetting
\citep{hu2026memoryagentbench,cheng2026amemgym,uddin2026memora}, and
memory-grounded action \citep{he2026memoryarena,zhao2026amabench,
wu2026longmemevalv2}. Recent designs span feedback-driven declarative and
procedural adaptation, joint personal and system memory, episode-level
knowledge versus execution demands, and hidden user-state recovery
\citep{ai2026memorybench,ma2026agentmemorybench,wang2026evomembench,
ma2026memprobe}. STALE, CAME-Bench, and Mem2ActBench target stale facts,
contextual applicability, and historical tool constraints
\citep{chao2026stale,yang2026came,shen2026mem2act}. These benchmarks measure
recall and downstream utility. \textsc{SciValidityBench} holds the focal record
fixed and varies the nonlocal conditions for its use, testing the
content--state--action relation.

\runinlabel{Scientific reasoning and scientific-agent benchmarks.}
Surveys chart scientific-agent stages, maturity, and reliability
\citep{wang2026hitchhiker}. Benchmarks cover knowledge and problem solving
\citep{wang2024scibench,laurent2024labbench}, data-driven discovery
\citep{majumder2025discoverybench,gu2024blade,chen2025scienceagentbench}, and
controlled or end-to-end research environments
\citep{bragg2026astabench,panigrahi2026heurekabench,kon2026expbench}.
Scientific-memory work treats full-paper recall as budgeted restoration, while
StatefulDiscovery externalizes investigation state for claim calibration
\citep{sheverev2026scientificmemory,chen2026statefuldiscovery}. These systems
score completion, restoration, or discovery quality. We fix the history and
action space, then test whether governing state permits evidence use.

\runinlabel{Provenance and scientific validity.}
PROV traces entities, activities, and agents \citep{moreau2013prov}, but
traceability alone does not establish admissibility. Preregistration fixes the
analysis plan before observation \citep{nosek2018preregistration}, while
selective inference accounts for data-dependent selection
\citep{fithian2014optimal}. Data leakage can also invalidate claims when
evaluation or artifact history is ignored \citep{kapoor2023leakage}. We turn
this principle into five mechanisms
and test whether readers reconstruct and obey the state governing valid use.

\runinlabel{Behavioral and counterfactual diagnosis.}
MemFail isolates summarization, storage, and retrieval failures
\citep{garg2026memfail}, while MemTrace follows memory evolution
\citep{deng2026memtrace}. CausalFlow and Causal Agent Replay intervene on
execution traces \citep{bonagiri2026causalflow,shah2026causalreplay}.
CheckList probes behavioral failures, whereas contrast
sets probe local decision boundaries \citep{ribeiro2020checklist,
gardner2020contrastsets}. SeekBench separates answers from epistemic process
\citep{shao2026seekbench}, and explicit state need not be mechanistically
faithful \citep{chen2026explicitstate}. Our interventions target scientific
state through one-value paired worlds, targeted corruption, and exact versus
edit-matched irrelevant repair.

\section{\textsc{SciValidityBench}}
\label{sec:benchmark-construction}

\subsection{Why scientific use requires global state}

A scientific notebook is an event trace, not yet a decision state. It may
faithfully record plans, runs, observations, access operations, and previous
choices while leaving their joint implication unstated. We call that
cross-record implication global scientific state: the collection of
constraints that determines which evidence may be combined, compared,
interpreted as missing, disclosed, or allowed to influence a prospective
action. ``Global'' refers to its scope across the history rather than to an
unbounded summary of everything that occurred.

This object separates three questions that are often conflated. Factuality
asks whether a recorded payload is correct. Provenance establishes where
the item came from and how it was produced. Use validity concerns what
that authentic record is licensed to support under the governing dependencies.
A verified measurement can therefore remain real and traceable while being
ineligible for pooling, confirmation, or resource allocation. The benchmark
targets this last relation without weakening the first two. The same payload
can therefore license different actions in otherwise paired histories.

Formally, let $E_{\mathrm{pub}}(w)$ denote the visible canonical-event history
of world $w$. A TaskSpec $\tau$ declares a deterministic compiler $C_\tau$,
which produces the decision-governing state
\begin{equation}
    G_\tau(w)=C_\tau\!\left(E_{\mathrm{pub}}(w)\right).
    \label{eq:state-compilation}
\end{equation}
It also provides a finite action space $\mathcal{A}_\tau$ and a programmatic
oracle $O_\tau(G_\tau(w))\in\mathcal{A}_\tau$. For an agent decision
$\widehat a$, scientific use validity is
\begin{equation}
    U(\widehat a,w,\tau)
    =\mathbf{1}\!\left[\widehat a=O_\tau(G_\tau(w))\right].
    \label{eq:use-validity}
\end{equation}
The measured object is consequently a state-conditioned action, rather than
recovery of an isolated span.

\subsection{What a TaskSpec specifies}

The basic unit of \textsc{SciValidityBench} is an executable TaskSpec,
not a fixed natural-language question. We write its functional contract as
\[
\tau=\langle D_\tau,W_\tau,C_\tau,\mathcal{A}_\tau,
O_\tau,I_\tau,S_\tau\rangle,
\]
where $D_\tau$ is a domain adapter, $W_\tau$ is a seeded world generator,
$C_\tau$ defines scientific-state compilation, $\mathcal{A}_\tau$ lists legal
decisions, $O_\tau$ returns the valid choice, $I_\tau$ constructs controlled
interventions, and $S_\tau$ scores responses and diagnostic relations. The
stored specification additionally fixes its version, episode schedule,
visibility rules, budgets, corruption operators, and data-split policy. One
template can thus instantiate many reproducible worlds while preserving the
same scientific rule. Appendix~\ref{app:catalog} summarizes catalog coverage,
contract fields, and one representative decision pattern from each module.

This contract turns an abstract validity concern into a complete test: the
generator creates the history, the compiler identifies what governs its use,
the model selects among declared actions, and code determines whether that
selection is legitimate. No free-form judge is needed for the primary outcome.
Stable identifiers and canonical serialization bind every score to the exact
events and state from which it was derived.

\runinlabel{Two orthogonal extension axes.}
The specification factorizes scientific surface from validity mechanism. To
add a domain, a developer supplies new entities, units, artifacts, and event
language through $D_\tau$ while reusing an existing state rule. To introduce a
new module, one declares its governing variables, compiler relations, oracle,
and interventions while retaining the common domain and scoring interfaces.
This separation makes the suite a benchmark factory rather than a collection
of unrelated datasets.

\subsection{How \textsc{SciStateForge} compiles scientific state}

\begin{figure*}[t]
\centering
\includegraphics[width=\textwidth]{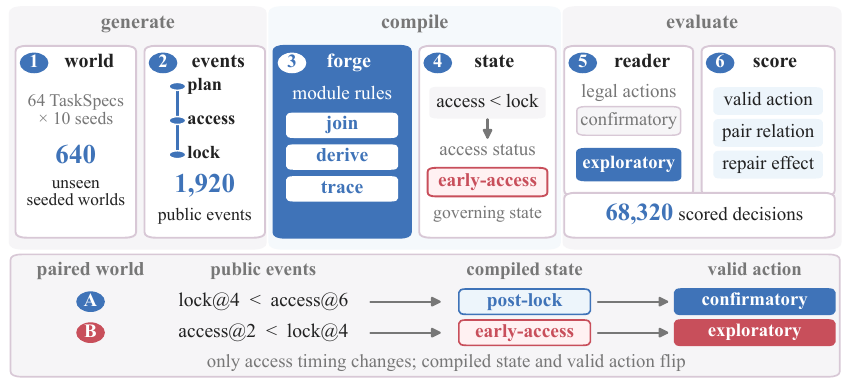}
\caption{\textbf{The \textsc{SciStateForge} pipeline.} A domain-specific
history is normalized, compiled into decision-governing state, rendered for a
model, and scored by an executable oracle. The worked pair changes one
scientific-state condition while keeping the task semantics fixed.}
\label{fig:scistateforge-pipeline}
\end{figure*}

\textsc{SciStateForge} realizes the TaskSpec contract as a compiler--reader
pipeline (Figure~\ref{fig:scistateforge-pipeline}). Its central operation is
not summarization: it derives explicit, source-linked state from dependencies
distributed across the trace.

\runinlabel{1. Instantiate a scientific world.}
Given a TaskSpec and seed, the domain adapter generates a multi-episode history
with a hidden world state and a reproducible public trajectory. Scientific
surfaces vary by domain, but every trajectory uses the same event vocabulary:
\textsc{Plan}, \textsc{Run}, \textsc{Observe}, \textsc{Access},
\textsc{Write}, \textsc{Retrieve}, and \textsc{Decide}.

\runinlabel{2. Canonicalize the history.}
Each event receives a stable identifier, actor, sequence position, referenced
objects, versioned payload, visibility, cost, and provenance. This normalized
layer provides byte-stable replay and prevents domain-specific formatting from
changing the underlying scientific world.

\runinlabel{3. Compile governing relations.}
Module reducers map public events into a Scientific State Graph (SSG). Its
nodes represent objects such as plans, observations, constructs, access
records, absences, costs, and decisions. Typed edges materialize relations
including \textsc{Selected\_From}, \textsc{Stopped\_After},
\textsc{Sealed\_Until}, \textsc{Comparable\_With},
\textsc{Absent\_Because}, and \textsc{Relevant\_To}. Derived fields retain
links to the events that justify them. They add no new observations while
making the scientific consequences of public evidence explicit.

This step is what distinguishes \textsc{SciStateForge} from a generic Event
Graph. The latter preserves what happened and connects events through sequence
or shared objects. The compiler additionally resolves what those occurrences
mean for the pending decision. Its flat SSG and typed Scientific Validity
Ledger renderings expose identical public nodes, attributes, and relations,
allowing representation to change without silently changing the compiled
content.

Figure~\ref{fig:scistateforge-pipeline} gives a concrete pass. The paired
worlds contain the same event types, but A records \texttt{lock@4} before
\texttt{access@6}, whereas B records \texttt{access@2} before
\texttt{lock@4}. The Blind Memory reducer joins those timestamps, derives
a \texttt{post-lock} or \texttt{early-access} status, and maps it to the
corresponding confirmatory or exploratory action. Each derived value remains
linked to the events that support it. The compiler terminates at this
source-linked state. The reader still selects from the legal actions, and the
TaskSpec oracle scores that selection. Compilation therefore resolves a
cross-event scientific dependency rather than retrieving a different fact or
executing the decision. Across the final catalog, 64 executable TaskSpecs crossed with ten unseen
seeds instantiate 640 seeded worlds and expose 1,920 public event records.

\runinlabel{4. Render, decide, and score.}
Oracle-only generator state remains with the validator. The reader receives a
projection derived from the public history together with the current request
and legal actions. Its choice is checked by the TaskSpec oracle, and the run
emits both an item-level validity score and the identifiers needed for paired
or repair-based diagnosis. Appendix~\ref{app:compiler} gives the canonical
event schema, deterministic passes, and exact memory-view definitions.

\subsection{Five modules across six scientific domains}

The catalog contains 64 versioned TaskSpecs spanning five validity modules and
six scientific domains. Table~\ref{tab:validity-modules} shows that the modules
are different mechanisms of one construct: whether the current scientific
state licenses the proposed use of a locally correct record.

\begin{table*}[t]
\caption{\textbf{Five modules, one use-validity construct.} Each module
isolates a distinct reason why a true record may or may not govern an action.}
\label{tab:validity-modules}
\centering
\small
\setlength{\tabcolsep}{3.5pt}
\begin{tabularx}{\textwidth}{@{}>{\raggedright\arraybackslash}p{2.1cm}>{\raggedright\arraybackslash}p{4.4cm}>{\raggedright\arraybackslash}X@{}}
\toprule
Module & Governing information & Invalid use under test \\
\midrule
Inferential Ledger & Search family, stopping rule, predeclared analysis scope, and
evidence consumption & Promoting an adaptively selected or repeatedly consumed
result to confirmation \\
Blind Memory & Plan lock, disclosure time, and authorized access roles &
Treating an outcome seen too early as an uncontaminated confirmatory analysis \\
Construct Drift & Metric definition, population, replicate meaning, and
version relation & Pooling measurements that share a name but are only
convertible or incommensurate \\
Absence Algebra & Attempt status, instrument outcome, censoring, and detection
mechanism & Collapsing not-run, failure, censoring, and confirmed zero into one
kind of blank \\
Sunk Cost & Prospective value and the relevance of historical investment &
Allowing irrecoverable past effort to alter a decision with identical future
payoffs \\
\bottomrule
\end{tabularx}
\end{table*}

Domains supply scientific settings rather than duplicate module definitions.
Machine learning contributes datasets, checkpoints, seeds, and metric
implementations; A/B testing uses variants, traffic slices, endpoints, and
scheduled looks; clinical research supplies protocols, cohorts, outcomes, and
censoring records. Bioinformatics introduces assays, samples, replicate
structure, and pipeline versions; systems research uses hardware, workloads,
runtime logs, and benchmark configurations; causal inference instantiates
exposures, estimands, adjustment sets, and identification choices. These
scientific surfaces are summarized in Table~\ref{tab:domain-surfaces}.

\begin{table*}[t]
\caption{\textbf{Six domain surfaces.} The adapter changes scientific objects
and language while the module retains the governing validity primitive.}
\label{tab:domain-surfaces}
\centering
\small
\setlength{\tabcolsep}{4pt}
\begin{tabularx}{\textwidth}{@{}>{\raggedright\arraybackslash}p{2.1cm}>{\raggedright\arraybackslash}X@{\hspace{12pt}}>{\raggedright\arraybackslash}p{2.1cm}>{\raggedright\arraybackslash}X@{}}
\toprule
Domain & Scientific surface & Domain & Scientific surface \\
\midrule
Machine learning & Datasets, checkpoints, seeds, metric implementations &
A/B testing & Variants, traffic slices, endpoints, scheduled looks \\
Clinical research & Protocols, cohorts, outcomes, censoring records &
Bioinformatics & Assays, samples, replicates, pipeline versions \\
Systems & Hardware, workloads, runtime logs, benchmark configurations &
Causal inference & Exposures, estimands, adjustment sets, identification choices \\
\bottomrule
\end{tabularx}
\end{table*}

This decomposition provides breadth without turning the suite into six
unrelated collections. Inferential Ledger, Construct Drift, and Absence
Algebra appear in every domain. Blind Memory and Sunk Cost are instantiated in
selected settings where their operational stories are most direct. Because world
generation is seeded, each TaskSpec yields controlled variation beneath the same
decision contract rather than repeating one memorized answer. Cross-domain
instances also resist surface shortcuts: one validity rule appears through
different entities and terminology, while one scientific setting hosts multiple
decision mechanisms.

\subsection{Interventions and programmatic diagnosis}

Correct actions alone cannot establish that a reader used the intended state.
SciValidityBench therefore builds diagnostic interventions into the benchmark
unit.

\runinlabel{State-changing counterfactuals.}
For each TaskSpec-defined pair, $I_\tau$ changes one decisive public value while
holding the TaskSpec, seed, surface form, legal actions, and non-target
information fixed. The contract declares whether the oracle relation must
\texttt{flip} or remain \texttt{invariant}. A model that repeats the same
choice in a flip pair may have succeeded on one side through answer preference
rather than state use.

\runinlabel{Corruption and matched repair.}
Graph operators can omit, unlink, relabel, merge, duplicate, or inject a
specific node or relation. Exact repair applies the inverse patch and restores
the target graph, including its canonical hash. An edit-matched irrelevant
repair leaves the decisive damage in place while modifying an unrelated
location with the same operation signature. Recovery unique to the exact
repair supports dependence on the intended scientific state. Equal recovery
from both repairs instead indicates a generic text or structure effect.

\runinlabel{Scoring beyond a leaderboard.}
The scorer records legal-action validity, the paired answer relation, and
repair behavior from deterministic contracts. These outputs distinguish four
useful failure patterns: failure to track the decisive value, a stable answer
bias, sensitivity to irrelevant history, and inability to exploit a restored
state. The benchmark consequently measures performance and supplies a causal
behavioral account of where that performance comes from.

Together, canonical events, declarative TaskSpecs, domain adapters, explicit
state compilation, and reversible interventions turn scientific use validity
into a reusable evaluation interface. The resulting pipeline can expand to a
new scientific setting without redefining the construct, or to a new validity
mechanism without rebuilding the surrounding evaluation stack. Every extension
retains the same diagnostic guarantees: executable scoring, state-changing
pairs, and repair at a known causal location. Breadth therefore does not trade
away failure interpretability.

\section{Experiments}
\label{sec:experiments}

Our evaluation follows a three-step evidence chain. We first measure whether
explicit scientific state improves valid decisions across independently
executed readers. Retrieval, input length, generic graph structure, and typed
serialization then test alternative explanations. Module effects,
counterfactual pairs, and targeted repair respectively locate the benefit,
verify state-dependent behavior, and diagnose failure.

\subsection{Experimental setup}

\runinlabel{Primary matrix.}
The main evaluation contains all 64 program-valid TaskSpecs. Ten unseen seeds
instantiate each specification for four open model families: Qwen3-14B,
Gemma-3-12B-IT,
Mistral-7B-Instruct-v0.3, and Phi-3.5-mini-instruct
\citep{qwen2025qwen3,gemma2025gemma3,jiang2023mistral,abdin2024phi3}.
Crossing those factors with five memory views produces
$64\times10\times4\times5=12{,}800$ original decisions. Flat SSG and typed
SVL additionally receive one state-changing companion world per
task--seed--reader, adding 5,120 counterfactual decisions. Completion is
outcome-independent: invalid JSON, illegal actions, and unparsable answers
count as invalid decisions.
Across the six frozen evaluation matrices reported in Appendix~\ref{app:protocol},
the final suite contains 68,320 scored decisions, of which 17,920 belong to the
primary matrix.

\runinlabel{Representations and controls.}
Raw Events exposes the full public sequence. A fixed BM25 retriever selects at most five
request-relevant event chunks. Event Graph preserves every public event and
object but connects them only through generic sequence and mention edges. The
two compiled conditions reveal the same public nodes, attributes, and governing
relations: Flat SSG serializes them as records, whereas SVL uses a typed ledger.
The prompt shell, current request, legal actions, seed, checkpoint, and
temperature-zero decoding remain fixed within each comparison. Models are
served with vLLM and return one action from the legal choice set
\citep{kwon2023vllm}.

\runinlabel{Outcomes and uncertainty.}
The primary measure is valid-action rate under the TaskSpec oracle. For every
representation contrast, we first average paired seed differences within a
TaskSpec and then weight all 64 specifications equally. Reported 95\%
intervals use 20,000 TaskSpec bootstrap resamples. Counterfactual relation
accuracy tests whether two answers obey the required flip or invariance. Repair
specificity compares restoring the damaged target with an edit-matched change
at an irrelevant location. These diagnostics complement the valid-action rate
by measuring how readers use the state behind each decision.
Appendix~\ref{app:protocol} details models, matrices, and estimands.

\subsection{Compiled state improves validity across settings}

\begin{figure}[t]
\centering
\includegraphics[width=\textwidth]{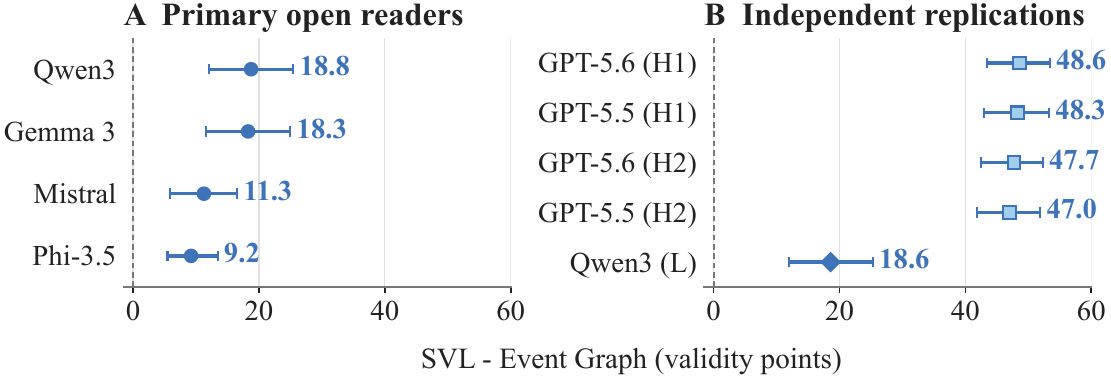}
\caption{\textbf{Main effect and independent replications.} Panel A reports
the TaskSpec-balanced primary contrast. Panel B shows the same contrast in two
hosted seed blocks (H1/H2) and a local Qwen3-14B replication (L).}
\label{fig:main-effect-replication}
\end{figure}

The primary matrix yields a consistent separation between scientific-state
compilation and generic event organization. As shown in
Table~\ref{tab:primary-results}, typed SVL exceeds Event Graph by 18.8, 18.3,
11.3, and 9.2 validity points for Qwen, Gemma, Mistral, and Phi, respectively.
All four 95\% intervals remain above zero. Figure~\ref{fig:main-effect-replication}A
summarizes the pattern. The result survives an exact-overlap audit. Removing
22 tasks leaves gains of 8.3--14.8 points for every reader across the remaining
42 tasks (Appendix~\ref{app:primary-results},
Table~\ref{tab:app-overlap-sensitivity}).

\begin{table}[b]
\caption{\textbf{Compiled state improves valid decisions across four model
families.} Each absolute-rate cell contains all 64 TaskSpecs at ten unseen
seeds. The final column first averages paired seed differences within each
TaskSpec and then weights TaskSpecs equally.}
\label{tab:primary-results}
\centering
\small
\setlength{\tabcolsep}{3.5pt}
\begin{tabular*}{\textwidth}{@{\extracolsep{\fill}}lcccccc@{}}
\toprule
Model & Raw & BM25 & Event & Flat & SVL & SVL$-$Event (95\% CI) \\
\midrule
Qwen3-14B       & 0.798 & 0.802 & 0.794 & \textbf{0.991} & 0.981 & \textbf{+0.188} [+0.120, +0.255] \\
Gemma-3-12B     & 0.733 & 0.719 & 0.719 & \textbf{0.903} & 0.902 & \textbf{+0.183} [+0.116, +0.250] \\
Mistral-7B          & 0.556 & 0.586 & 0.566 & \textbf{0.692} & 0.678 & \textbf{+0.113} [+0.059, +0.166] \\
Phi-3.5-mini    & 0.480 & 0.475 & 0.478 & \textbf{0.609} & 0.570 & \textbf{+0.092} [+0.055, +0.134] \\
\bottomrule
\end{tabular*}
\end{table}

Event Graph stays close to Raw Events despite containing all public events and
their basic relations. Compilation materializes consequences such as
\textsc{Comparable\_With}, \textsc{Absent\_Because}, or
\textsc{Stopped\_After}. The reader no longer has to reconstruct those
cross-record dependencies from sequence alone. The 9.2--18.8 point gain comes
from exposing the scientific relations that govern the current decision, not
from organizing the same event history as a generic graph.

Flat SSG isolates compiled content from ledger syntax. It reaches
0.991/0.903/0.692/0.609 across the four readers, slightly above typed SVL in
every row. Both renderings disclose identical nodes, attributes, and governing
relations, so their shared improvement comes from the compiled scientific
state. The small Flat--SVL differences show that readers vary in how they
consume the same state.

Independent executions reproduce the contrast in
Figure~\ref{fig:main-effect-replication}B. Hosted runs yield 47.0--48.6 point
advantages across two readers and two seed blocks, with positive intervals
throughout. An independent Qwen replication produces an 18.6-point gain
$[12.0,25.3]$, while Flat and typed SVL remain within five points. The direction
persists across model identities and serving stacks. Appendices~\ref{app:primary-results}
and~\ref{app:replications} give full results and replications.

\subsection{What drives the gain?}

Retrieval does not reproduce the cross-family pattern. BM25 raises Mistral by
3.0 points and Qwen by 0.3, but lowers the other two readers. Event Graph is the
longest view for every tokenizer yet remains close to Raw Events. The shorter
compiled projections lead instead, tying performance to explicit governing
relations rather than retrieval depth or context length.

The source-level matrix changes one canonical public value before constructing
any view. Matched Source exposes the complete derived-free fact closure and
exactly matches Flat SSG's full-chat token count. Across 12 balanced design
families, Flat SSG exceeds Event Graph for every reader
(Table~\ref{tab:causal-controls}). Scientific-relation compilation remains
beneficial after the shared upstream intervention.

\begin{table}[b]
\caption{\textbf{Source- and serialization-level controls reveal
reader-specific mechanisms.} Effects are balanced over 12 design
families, with 95\% family-resampling intervals. Matched Source and Flat SSG
have exactly equalized full-chat token counts.}
\label{tab:causal-controls}
\centering
\small
\setlength{\tabcolsep}{3.5pt}
\begin{tabular*}{\textwidth}{@{\extracolsep{\fill}}lccc@{}}
\toprule
Model & Flat$-$Event Graph & Flat$-$Matched Source & SVL$-$Flat \\
\midrule
Qwen3-14B      & +0.174 [+0.056, +0.300] & +0.194 [+0.066, +0.335] & +0.003 [+0.000, +0.010] \\
Gemma-3-12B    & +0.133 [+0.000, +0.273] & +0.011 [$-$0.136, +0.162] & +0.240 [+0.125, +0.356] \\
Mistral-7B          & +0.207 [+0.078, +0.347] & +0.193 [+0.064, +0.333] & $-$0.005 [$-$0.023, +0.009] \\
Phi-3.5-mini   & +0.189 [+0.045, +0.375] & $-$0.295 [$-$0.460, $-$0.099] & +0.488 [+0.276, +0.696] \\
\bottomrule
\end{tabular*}
\end{table}

Exact token matching separates compilation from presentation. Flat outperforms
Matched Source by 19.4 points for Qwen and 19.3 for Mistral, changes Gemma by
1.1, and trails it by 29.5 for Phi. Typed SVL is effectively tied with Flat for
Qwen and Mistral but improves Gemma by 24.0 points and Phi by 48.8. The common
advantage lies in compiled state. The easiest serialization depends on the
reader.

\subsection{Where compilation helps and whether readers use state}

Benefits concentrate in particular module--reader combinations
(Figure~\ref{fig:state-use-diagnosis}A). Blind Memory rises by 21.3--52.5 points
across the four readers, and Inferential Ledger improves in every family.
Absence Algebra primarily benefits Mistral. Sunk Cost is unchanged for Mistral
and Phi but lower for Qwen and Gemma. The aggregate effect therefore reflects
structured heterogeneity rather than a uniform catalog-wide lift.

Module effects locate where compilation helps. Paired worlds test whether the
decisive state actually governs behavior. Counterfactual pairs hold task
surface, graph structure, legal choices, and non-target information fixed while
changing one public value. Relation accuracy scores the oracle-declared flip or
invariance, so a fortunate answer on one side does not count as state tracking.

Panel B reveals a sharper hierarchy than
single-world accuracy. For Flat SSG/typed SVL, relation accuracy reaches
1.000/0.998 for Qwen and 0.906/0.905 for Gemma, then falls to 0.561/0.550 for
Mistral and 0.436/0.361 for Phi. Stronger readers convert explicit state into
counterfactually consistent behavior. Weaker ones often retain stable answer
preferences.

Panel C supplies an independent localization test. Exact restoration exceeds
an edit-matched irrelevant repair by 26.9
points for Qwen, 32.8 for Gemma, and 15.0 for Mistral, but changes Phi by
$-$0.3 points. The four-reader mean is 18.6. Cross-domain corruption
fingerprints are stable for five of six operators in Qwen and Gemma, versus two
for Mistral and one for Phi. High relation accuracy and stable fingerprints
coincide in the stronger readers, whereas Mistral and Phi show weaker tracking
and less specific recovery. The contrast distinguishes damage to a governing
relation from a broader failure to use compiled state. Appendices~\ref{app:counterfactuals}
and~\ref{app:repair} give complete diagnostic results.

\begingroup
\setlength{\intextsep}{0pt}
\begin{figure}[!ht]
\vspace*{12pt}
\centering
\includegraphics[width=\textwidth]{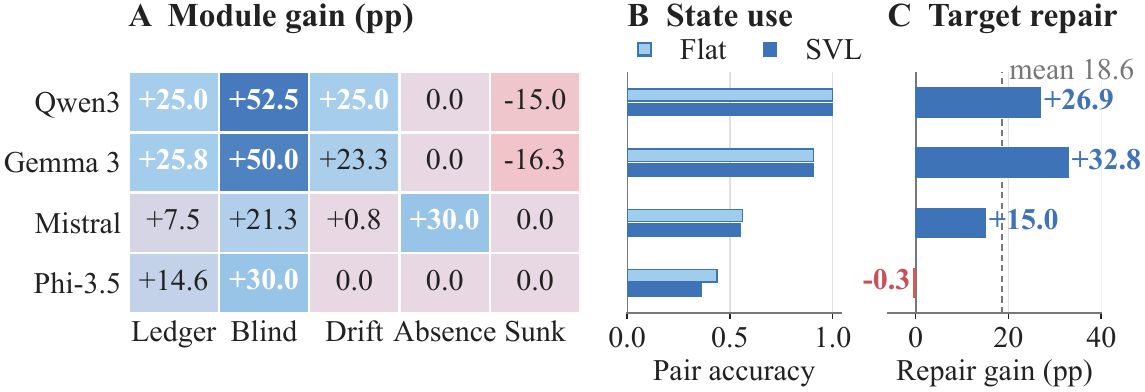}
\caption{\textbf{State-use and failure diagnosis.} Panel A reports module-level
validity gains (SVL$-$Event Graph, percentage points) and shows where compilation
helps. Value pairs test whether behavior follows a changed
state. Target and matched repairs distinguish restoration of the governing
information from a generic edit effect. All panels use a shared reader order
and Panel A labels every module--reader contrast.}
\label{fig:state-use-diagnosis}
\end{figure}
\endgroup

\section{Conclusion}
\label{sec:conclusion}

Scientific memory must recover a true record and determine whether it is
licensed for the current decision. \textsc{SciValidityBench} makes this
scientific-use validity executable through 64 TaskSpecs spanning five mechanisms
and six domains, while \textsc{SciStateForge} compiles distributed histories
into source-linked decision state. Across four model families, compiled state
improves valid-action selection over Event Graph by 9.2--18.8 points. The
advantage persists after removing exact state--action overlaps and cannot be
reproduced by retrieval, longer inputs, or generic graph structure.
Counterfactual pairs and matched repairs distinguish genuine state dependence
from answer preference and localize failures, shifting memory evaluation from
retaining facts to reconstructing the conditions under which they may be
compared, promoted, or acted upon. TaskSpecs separate scientific surfaces from
validity rules for agents writing laboratory histories.

\clearpage
\subsection*{AI use statement}

Generative AI tools were used only to polish the manuscript's language,
including grammar, clarity, and style. They did not determine the research
ideas, methods, experimental design, implementation, data, analysis, or
conclusions. The authors reviewed and verified the complete manuscript and
take full responsibility for all of its content.

\subsection*{Ethics Statement}

The benchmark uses procedurally generated histories and fictitious clinical
scenarios. It involves no human participants, patient records, personal data,
or confidential information. No human-subjects research was conducted, so IRB
review was not required. The authors take responsibility for the ethical use
and release of the benchmark, experiments, and associated materials.

\subsection*{Reproducibility statement}

The appendices document the TaskSpec contracts, memory-view definitions,
evaluation settings, programmatic scoring, and diagnostic interventions. The
\href{https://github.com/SciValidityBench/SciValidityBench}{public GitHub repository} provides the final 64-TaskSpec catalog,
deterministic generators spanning six domains, the SciStateForge compiler, all
five evaluated memory views, counterfactual and corruption/repair operators,
executable examples, and release tests. Every benchmark world is generated
deterministically from a TaskSpec and seed, and oracle-only state is excluded
from model-visible inputs.

\bibliography{references}
\bibliographystyle{arxiv_preprint}

\clearpage
\appendix
\section{Benchmark Design Details}
\label{app:catalog}

\subsection{Catalog coverage}

The benchmark contains 64 executable TaskSpecs spanning five scientific-use
validity modules and six domains.  Modules define why a record may or may not
govern a decision.  Domains determine the scientific objects, units, and event
language used to express that rule.  This factorization allows either axis to
expand without redesigning the evaluation interface.  The resulting coverage
is summarized in Table~\ref{tab:app-catalog-coverage}.

\begin{table}[!htbp]
\caption{\textbf{Module--domain coverage of the 64 TaskSpecs.}  Inferential
Ledger, Construct Drift, and Absence Algebra cover all six domains.  Blind
Memory and Sunk Cost use domains where their interventions have a direct
operational interpretation.}
\label{tab:app-catalog-coverage}
\centering
\small
\begin{tabular}{@{}lrrrrrrr@{}}
\toprule
Module & ML & A/B & Clinical & Bioinfo. & Systems & Causal & Total \\
\midrule
Inferential & 4 & 4 & 4 & 4 & 4 & 4 & 24 \\
Blind & 2 & 2 & 2 & 0 & 0 & 2 & 8 \\
Construct & 2 & 2 & 2 & 2 & 2 & 2 & 12 \\
Absence & 2 & 2 & 2 & 2 & 2 & 2 & 12 \\
Sunk Cost & 2 & 2 & 0 & 2 & 2 & 0 & 8 \\
\midrule
Total & 12 & 12 & 10 & 10 & 10 & 10 & 64 \\
\bottomrule
\end{tabular}

\end{table}

\subsection{Executable TaskSpec contract}

A TaskSpec is a program that produces a reproducible family of scientific
worlds, not a fixed natural-language question.  It contains every component
needed to generate a history, compile the decision-governing state, intervene
on that state, and score the resulting action.  Table~\ref{tab:app-taskspec-contract}
specifies the shared executable interface.

\begin{table}[!htbp]
\caption{\textbf{Executable TaskSpec contract.}  The components map a generated
history to compiled state, controlled interventions, valid actions, and
diagnostic scores.}
\label{tab:app-taskspec-contract}
\centering
\small
\begin{tabularx}{\textwidth}{@{}p{2.5cm}p{3.4cm}X@{}}
\toprule
Component & Output & Function \\
\midrule
Domain adapter $D_\tau$ & Entities, units, and event language & Changes the
scientific surface while preserving the validity primitive \\
World generator $W_\tau$ & Hidden state and public history & Uses a declared
seed and version to produce reproducible event sequences \\
State compiler $C_\tau$ & Public nodes, attributes, and relations & Derives
decision-governing state only from model-visible records \\
Action space $\mathcal A_\tau$ & Finite legal decisions & Fixes the response
contract before evaluation \\
Oracle $O_\tau$ & Valid action & Computes the answer from the governing state
without a free-form model judge \\
Intervention $I_\tau$ & Counterfactual and corrupted worlds & Changes a
decisive value or state component under a declared relation \\
Scorer $S_\tau$ & Validity and diagnostic outcomes & Measures action validity,
pair consistency, and repair specificity \\
\bottomrule
\end{tabularx}
\end{table}

The stored specification also fixes its schema version, episode schedule,
visibility policy, legal choices, prompt surface, corruption operators, and
data split.  Before model execution, validators reject a counterfactual whose
oracle relation is inconsistent and a repair that does not reconstruct the
target state.

\subsection{Representative module examples}

Table~\ref{tab:app-worked-examples} instantiates the common interface once per
module, highlighting the state relation that changes or preserves the valid
decision.

\begin{table}[!htbp]
\caption{\textbf{One decision pattern per validity module.}  Each example
uses the same generator--compiler--oracle interface while changing the state
relation that licenses the action.}
\label{tab:app-worked-examples}
\centering
\small
\begin{tabularx}{\textwidth}{@{}p{2.2cm}p{5.2cm}X@{}}
\toprule
Module & Decisive scientific state & Decision change \\
\midrule
Inferential Ledger & Whether the result followed the declared search family
and stopping rule & Confirmation becomes exploratory after adaptive selection \\
Blind Memory & Whether outcome access occurred before the analysis plan was
locked & A confirmatory analysis becomes compromised after premature access \\
Construct Drift & Whether two identically named measurements share a
comparable definition & Results may be pooled only when their constructs are
equivalent or validly converted \\
Absence Algebra & Whether a missing record means not run, failure, censoring,
or a true zero & The next action changes among initiating, repeating,
censor-aware analysis, and recording zero \\
Sunk Cost & Whether historical investment is relevant to prospective value &
The preferred project remains unchanged when only irrecoverable past effort
changes \\
\bottomrule
\end{tabularx}
\end{table}

\section{SciStateForge Compilation and Memory Views}
\label{app:compiler}

\subsection{Canonical event layer}

SciStateForge first converts domain-specific histories into a shared event
layer.  Domain payloads remain versioned data rather than being embedded in
free-form notebook prose.  Table~\ref{tab:app-event-schema} lists the normalized
fields that every domain supplies.

\begin{table}[!htbp]
\caption{\textbf{Canonical event fields.}  Normalization makes domain
histories replayable before scientific-state relations are derived.}
\label{tab:app-event-schema}
\centering
\small
\begin{tabularx}{\textwidth}{@{}p{2.8cm}X@{}}
\toprule
Field & Role \\
\midrule
Stable identifiers & Bind the world, episode, event, actor, and referenced objects \\
Sequence and visibility & Determine event order and when a record becomes public \\
Versioned payload & Store domain values under a declared schema \\
Cost and provenance & Retain resource use and the source that produced the event \\
Event type & Normalize \textsc{Plan}, \textsc{Run}, \textsc{Observe},
\textsc{Access}, \textsc{Write}, \textsc{Retrieve}, and \textsc{Decide} \\
\bottomrule
\end{tabularx}
\end{table}

\subsection{Deterministic compilation}

For a world $w$ and TaskSpec $\tau$, the compiler performs four passes:

\begin{list}{\arabic{enumi}.}{%
    \usecounter{enumi}%
    \setlength{\leftmargin}{1.4em}%
    \setlength{\labelwidth}{1em}%
    \setlength{\labelsep}{0.4em}%
    \setlength{\itemindent}{0pt}%
    \setlength{\listparindent}{0pt}%
    \setlength{\topsep}{0pt}%
    \setlength{\partopsep}{0pt}%
    \setlength{\parsep}{0pt}%
    \setlength{\itemsep}{0pt}%
    \renewcommand{\makelabel}[1]{#1\hfil}%
}
    \item validate and order the public canonical events
    $E_{\mathrm{pub}}(w)$;
    \item instantiate typed objects for plans, observations, constructs,
    access records, absences, costs, and pending decisions;
    \item derive governing relations such as \textsc{Selected\_From},
    \textsc{Stopped\_After}, \textsc{Sealed\_Until},
    \textsc{Comparable\_With}, \textsc{Absent\_Because}, and
    \textsc{Relevant\_To};
    \item project public nodes and relations into a model-facing view while
    retaining links from derived fields to their supporting events.
\end{list}

The resulting state $G_\tau(w)=C_\tau(E_{\mathrm{pub}}(w))$ is a deterministic
closure over the visible history.  It introduces no new observation: the
compiler makes cross-record consequences explicit while keeping oracle-only
generator state outside the reader input.

\subsection{Memory-view definitions}

The five views in Table~\ref{tab:app-memory-views} progressively separate raw
history, retrieval, generic event structure, and compiled scientific state.

\begin{table}[!htbp]
\caption{\textbf{Five primary memory views.}  Flat SSG and typed SVL expose
the same compiled nodes, attributes, and relations.  Event Graph preserves all
source events but omits scientific-state derivation.}
\label{tab:app-memory-views}
\centering
\small
\begin{tabularx}{\textwidth}{@{}p{2.2cm}p{4.3cm}X@{}}
\toprule
View & Model-visible content & Controlled question \\
\midrule
Raw Events & Complete ordered public event sequence & Can the reader derive
the valid action directly from the full history? \\
BM25 Events & Top five request-relevant event chunks & Does conventional
retrieval recover the governing evidence? \\
Event Graph & All public events and objects with generic sequence and mention
edges & Does organizing events as a graph recover use validity? \\
Flat SSG & Every public compiled node, attribute, and relation as flat records
& What changes when decision-governing state is explicit without graph syntax? \\
Typed SVL & The same compiled content as typed nodes and relations & How does
a structured scientific ledger affect different readers? \\
\bottomrule
\end{tabularx}
\end{table}

Input length does not track representation quality: the median counts in
Table~\ref{tab:app-primary-tokens} shows that Event Graph is the longest view
for every reader.

\begin{table}[!htbp]
\caption{\textbf{Median input tokens in the primary representation matrix.}
Each model--view cell contains 640 prompts.  Token counts use the reader's own
tokenizer.  Event Graph is longest for every model.}
\label{tab:app-primary-tokens}
\centering
\small
\begin{tabular}{@{}lrrrrr@{}}
\toprule
Model & Raw & BM25 & Event Graph & Flat SSG & SVL \\
\midrule
Qwen3-14B & 631 & 718 & 1,044 & 460 & 439 \\
Gemma-3-12B & 678 & 775 & 1,112 & 498 & 468 \\
Mistral-7B & 790 & 902 & 1,326 & 598 & 568 \\
Phi-3.5-mini & 798 & 909 & 1,344 & 613 & 580 \\
\bottomrule
\end{tabular}

\end{table}

\section{Experimental Details}
\label{app:protocol}

\subsection{Readers and evaluation matrices}

The open-weight evaluation uses Qwen3-14B, Gemma-3-12B-IT,
Mistral-7B-Instruct-v0.3, and Phi-3.5-mini-instruct.  Each model is served alone
through vLLM with deterministic decoding and the same choice-only output
contract.  Invalid JSON, illegal choices, refusals, and unparsable responses
remain invalid decisions rather than being resampled.

Table~\ref{tab:app-matrices} connects each evaluation matrix to its scientific
question and reports the number of model decisions.

\begin{table}[!htbp]
\caption{\textbf{Evaluation matrices.}  Each matrix targets a distinct
question: representation quality, state use, failure localization,
generalization, or source-level control.}
\label{tab:app-matrices}
\centering
\small
\setlength{\tabcolsep}{3pt}
\begin{tabularx}{\textwidth}{@{}p{2.9cm}rrrp{4.2cm}r@{}}
\toprule
Purpose & Tasks & Seeds & Readers & Conditions & Decisions \\
\midrule
Primary representations and value pairs & 64 & 10 & 4 & Five original
views plus companion worlds for the two compiled views & 17,920 \\
Corruption and repair & 24 & 6 & 4 & Intact, corrupted, exact repair, and
matched irrelevant repair & 4,896 \\
Additional memory policies & 24 & 10 & 4 & Seven memory conditions & 6,720 \\
Independent hosted replication & 64 & Two blocks & 2 & Primary representation
contrasts & 17,920 \\
Independent Qwen replication & 64 & Matched & 1 & Original views and value
pairs & 4,480 \\
Source and serialization controls & 64 & 4 & 4 & Eight views over two source
worlds & 16,384 \\
\bottomrule
\end{tabularx}
\end{table}

\subsection{Estimands and uncertainty}

For views $a$ and $b$, the analysis averages paired seed differences within
TaskSpec $t$,
\[
    \bar d_{m,t}=\frac{1}{|S|}\sum_{s\in S}
    \left(U_{m,t,s,a}-U_{m,t,s,b}\right),
\]
and then weights TaskSpecs equally.  The reported 95\% intervals use 20,000
TaskSpec resamples.  Family-balanced controls instead weight the 12
module--design families equally and resample those families.

For counterfactual pairs, relation accuracy tests whether the reader's two
actions obey the required flip or invariance.  Both-world validity requires
the oracle action on each side.  Repair specificity subtracts recovery after
an edit-matched irrelevant change from recovery after restoring the damaged
target.

\subsection{Scoring and completion}

All program-valid tasks in a scheduled matrix enter its denominator.  The
scorer records action validity, parse status, pair relation, TaskSpec version,
and the model-visible input.  Infrastructure retries do not replace a terminal
model response, preventing the evaluation from retaining only easy-to-parse or
successful outputs.

\section{Complete Representation Results}
\label{app:primary-results}

\subsection{Absolute rates and paired contrasts}

Compiled views lead Event Graph for every open reader.  Flat SSG is slightly
higher than typed SVL in this matrix, indicating that their shared scientific
state drives the cross-family improvement.  Table~\ref{tab:app-condition-rates}
reports absolute validity.  Table~\ref{tab:app-primary-contrasts} gives the
paired representation effects.

\begin{table}[!htbp]
\caption{\textbf{Original-world validity over all 64 TaskSpecs and ten seeds.}
Each model--view cell contains 640 decisions.}
\label{tab:app-condition-rates}
\centering
\small
\begin{tabular}{@{}lrrrrr@{}}
\toprule
Model & Raw & BM25 & Event Graph & Flat SSG & SVL \\
\midrule
Qwen3-14B & 0.798 & 0.802 & 0.794 & 0.991 & 0.981 \\
Gemma-3-12B & 0.733 & 0.719 & 0.719 & 0.903 & 0.902 \\
Mistral-7B & 0.556 & 0.586 & 0.566 & 0.692 & 0.678 \\
Phi-3.5-mini & 0.480 & 0.475 & 0.478 & 0.609 & 0.570 \\
\bottomrule
\end{tabular}

\end{table}

\begin{table}[!htbp]
\caption{\textbf{TaskSpec-balanced representation contrasts.}  Intervals
come from 20,000 TaskSpec resamples.  Retrieval does not reproduce the
cross-family compiled-state advantage.}
\label{tab:app-primary-contrasts}
\centering
\small
\begin{tabular*}{\textwidth}{@{\extracolsep{\fill}}lrr@{}}
\toprule
Model & BM25$-$Raw & Event$-$Raw \\
\midrule
Qwen3-14B & +0.003 [-0.003, +0.011] & -0.005 [-0.013, +0.000] \\
Gemma-3-12B & -0.014 [-0.031, -0.002] & -0.014 [-0.033, +0.002] \\
Mistral-7B & +0.030 [+0.005, +0.059] & +0.009 [-0.023, +0.044] \\
Phi-3.5-mini & -0.005 [-0.016, +0.003] & -0.002 [-0.011, +0.006] \\
\bottomrule
\end{tabular*}

\begin{tabular*}{\textwidth}{@{\extracolsep{\fill}}lrr@{}}
\toprule
Model & SVL$-$Event & SVL$-$Flat \\
\midrule
Qwen3-14B & +0.188 [+0.120, +0.255] & -0.009 [-0.023, +0.000] \\
Gemma-3-12B & +0.183 [+0.116, +0.250] & -0.002 [-0.005, +0.000] \\
Mistral-7B & +0.113 [+0.059, +0.166] & -0.014 [-0.055, +0.028] \\
Phi-3.5-mini & +0.092 [+0.055, +0.134] & -0.039 [-0.070, -0.014] \\
\bottomrule
\end{tabular*}

\end{table}

\subsection{Module and domain structure}

Blind Memory and Inferential Ledger consistently reward explicit governing
relations.  Sunk Cost instead tests whether a reader can suppress an irrelevant
historical investment.  Exposing more structured history does not guarantee an
improvement when the reader remains attracted to that distractor.  The complete
reader--module matrix appears in Table~\ref{tab:app-module-rates}.

\begin{table}[!htbp]
\caption{\textbf{Validity by reader, module, and representation.}  Each cell
uses ten seeds and every primary TaskSpec in the corresponding module.}
\label{tab:app-module-rates}
\centering
\small
\setlength{\tabcolsep}{4pt}
\begin{tabular}{@{}llrrrrr@{}}
\toprule
Model & Module & Raw & BM25 & Event & Flat & SVL \\
\midrule
Qwen3-14B & Inferential & 0.750 & 0.750 & 0.750 & 1.000 & 1.000 \\
Qwen3-14B & Blind & 0.512 & 0.512 & 0.475 & 1.000 & 1.000 \\
Qwen3-14B & Construct & 0.750 & 0.767 & 0.750 & 1.000 & 1.000 \\
Qwen3-14B & Absence & 1.000 & 1.000 & 1.000 & 1.000 & 1.000 \\
Qwen3-14B & Sunk Cost & 1.000 & 1.000 & 1.000 & 0.925 & 0.850 \\
\addlinespace[2pt]
Gemma-3-12B & Inferential & 0.633 & 0.625 & 0.617 & 0.875 & 0.875 \\
Gemma-3-12B & Blind & 0.500 & 0.500 & 0.500 & 1.000 & 1.000 \\
Gemma-3-12B & Construct & 0.808 & 0.750 & 0.767 & 1.000 & 1.000 \\
Gemma-3-12B & Absence & 1.000 & 1.000 & 1.000 & 1.000 & 1.000 \\
Gemma-3-12B & Sunk Cost & 0.750 & 0.750 & 0.750 & 0.600 & 0.588 \\
\addlinespace[2pt]
Mistral-7B & Inferential & 0.500 & 0.500 & 0.554 & 0.704 & 0.629 \\
Mistral-7B & Blind & 0.500 & 0.500 & 0.500 & 0.512 & 0.713 \\
Mistral-7B & Construct & 0.750 & 0.750 & 0.742 & 0.750 & 0.750 \\
Mistral-7B & Absence & 0.550 & 0.708 & 0.500 & 0.842 & 0.800 \\
Mistral-7B & Sunk Cost & 0.500 & 0.500 & 0.500 & 0.525 & 0.500 \\
\addlinespace[2pt]
Phi-3.5-mini & Inferential & 0.496 & 0.500 & 0.487 & 0.650 & 0.633 \\
Phi-3.5-mini & Blind & 0.500 & 0.500 & 0.512 & 1.000 & 0.812 \\
Phi-3.5-mini & Construct & 0.600 & 0.567 & 0.600 & 0.600 & 0.600 \\
Phi-3.5-mini & Absence & 0.300 & 0.300 & 0.300 & 0.350 & 0.300 \\
Phi-3.5-mini & Sunk Cost & 0.500 & 0.500 & 0.500 & 0.500 & 0.500 \\
\bottomrule
\end{tabular}

\end{table}

The typed-SVL minus Event Graph contrast is positive in all 24
reader--domain cells.  Clinical and causal cells produce the largest gains for
Qwen and Gemma, while Mistral peaks in bioinformatics.  These descriptive
differences partly reflect the catalog's unequal module mix and do not isolate
a causal domain effect.  Phi's effects are smaller but remain positive across the six domains
(Table~\ref{tab:app-domain-effects}).

\begin{table}[!htbp]
\caption{\textbf{Typed SVL minus Event Graph by scientific domain.}
TaskSpecs are equally weighted within each reader--domain cell.}
\label{tab:app-domain-effects}
\centering
\small
\begin{tabular}{@{}lrrrrrr@{}}
\toprule
Model & ML & A/B & Clinical & Bioinfo. & Systems & Causal \\
\midrule
Qwen3-14B & +0.183 & +0.183 & +0.250 & +0.120 & +0.120 & +0.270 \\
Gemma-3-12B & +0.192 & +0.167 & +0.260 & +0.110 & +0.120 & +0.250 \\
Mistral-7B & +0.100 & +0.100 & +0.090 & +0.190 & +0.070 & +0.130 \\
Phi-3.5-mini & +0.092 & +0.108 & +0.110 & +0.060 & +0.070 & +0.110 \\
\bottomrule
\end{tabular}

\end{table}

\subsection{Dependence and shortcut sensitivities}

Equal family weighting and leave-one-family-out analysis retain positive
effects for all readers.  This prevents a large template family from
dominating the catalog-wide estimate.  Table~\ref{tab:app-family-sensitivity}
reports both estimates.

\begin{table}[!htbp]
\caption{\textbf{Sensitivity to design-family dependence.}  Effects weight
12 module--design families equally.  The final column is the range obtained by
leaving out one family at a time.  LOGO denotes leave-one-family-out.}
\label{tab:app-family-sensitivity}
\centering
\small
\begin{tabular}{@{}lrrrrr@{}}
\toprule
Model & Families & Positive & Effect & 95\% interval & Family LOGO \\
\midrule
Qwen3-14B & 12 & 5 & +0.188 & [+0.033, +0.342] & [+0.155, +0.232] \\
Gemma-3-12B & 12 & 6 & +0.181 & [+0.025, +0.336] & [+0.151, +0.227] \\
Mistral-7B & 12 & 6 & +0.112 & [+0.017, +0.217] & [+0.083, +0.134] \\
Phi-3.5-mini & 12 & 4 & +0.099 & [+0.021, +0.194] & [+0.064, +0.108] \\
\bottomrule
\end{tabular}

\end{table}

Some TaskSpecs contain a public state value whose literal string equals a legal
choice.  Removing all 22 such tasks leaves 42 tasks without this equality.
Every reader retains a positive typed-SVL minus Event Graph effect, ruling out
direct value copying as the explanation for the main pattern
(Table~\ref{tab:app-overlap-sensitivity}).

\begin{table}[!htbp]
\caption{\textbf{Exact-choice-overlap sensitivity.}  The primary contrast is
recomputed after removing all tasks with literal equality between a compiled
state value and a legal action.  LOTO denotes leave-one-task-out.}
\label{tab:app-overlap-sensitivity}
\centering
\small
\begin{tabular}{@{}lrrrrr@{}}
\toprule
Model & Tasks & Positive & Effect & 95\% interval & LOTO range \\
\midrule
Qwen3-14B & 42 & 12 & +0.143 & [+0.083, +0.214] & [+0.134, +0.146] \\
Gemma-3-12B & 42 & 13 & +0.148 & [+0.081, +0.219] & [+0.137, +0.151] \\
Mistral-7B & 42 & 22 & +0.131 & [+0.057, +0.202] & [+0.122, +0.146] \\
Phi-3.5-mini & 42 & 12 & +0.083 & [+0.043, +0.126] & [+0.076, +0.085] \\
\bottomrule
\end{tabular}

\end{table}

\section{Counterfactual State-Use Evaluation}
\label{app:counterfactuals}

\subsection{Pair construction}

A companion world changes one decisive public value while preserving the
TaskSpec, seed, request, legal choices, surface language, and non-target
information.  The oracle states whether the correct action must flip or remain
invariant.  Relation accuracy is one only when the reader's two actions obey
that requirement.

This test is stricter than scoring the worlds separately.  Repeating
\texttt{exploratory}, for example, may be correct on the contaminated side of
an Inferential Ledger pair but cannot count as state tracking when the clean
side requires \texttt{confirm}.  Conversely, a Sunk Cost pair can change past
investment while requiring an invariant action.

\subsection{Aggregate and module-level results}

Table~\ref{tab:app-pair-summary} compares single-world validity, both-world
validity, and relation accuracy for the two compiled views.

\begin{table}[!htbp]
\caption{\textbf{Aggregate state-use metrics over 640 pairs per
reader--compiled-view cell.}  Both-valid requires correct actions on both
sides.  Relation accuracy also recognizes valid invariance.}
\label{tab:app-pair-summary}
\centering
\small
\begin{tabular}{@{}lrrrr@{}}
\toprule
Model & Flat relation acc. & SVL relation acc. & Flat both-valid & SVL both-valid \\
\midrule
Qwen3-14B & 1.000 & 0.998 & 0.991 & 0.981 \\
Gemma-3-12B & 0.906 & 0.905 & 0.856 & 0.855 \\
Mistral-7B & 0.561 & 0.550 & 0.442 & 0.412 \\
Phi-3.5-mini & 0.436 & 0.361 & 0.327 & 0.270 \\
\bottomrule
\end{tabular}

\end{table}

Qwen follows the required relation almost perfectly in both compiled views,
and Gemma remains above 0.90.  Mistral and Phi show a larger gap between
single-world validity and state-dependent consistency.  The module breakdown
localizes these differences: Mistral benefits from the typed Blind
representation but remains weak on Inferential pairs.  Phi's Sunk Cost relation
accuracy is 1.000 in both views, but both-world validity is only 0.500, so this
invariant-pair score alone does not establish correct state use.  Phi also fails
most Absence pairs
(Table~\ref{tab:app-pair-modules}).

\begin{table}[!htbp]
\caption{\textbf{Counterfactual relation accuracy and both-world validity by
module.}  Each cell reports Flat/SVL.  Module sizes are
240/80/120/120/80 pairs in the displayed order.}
\label{tab:app-pair-modules}
\centering
\small
\setlength{\tabcolsep}{3pt}
\begin{tabular*}{\textwidth}{@{\extracolsep{\fill}}lccccc@{}}
\toprule
Model & Infer. & Blind & Construct & Absence & Sunk \\
\midrule
\multicolumn{6}{@{}l}{Relation accuracy (Flat/SVL)} \\
Qwen3-14B & 1.000/1.000 & 1.000/1.000 & 1.000/1.000 & 1.000/1.000 & 1.000/0.988 \\
Gemma-3-12B & 0.750/0.750 & 1.000/1.000 & 1.000/1.000 & 1.000/1.000 & 1.000/0.988 \\
Mistral-7B & 0.408/0.263 & 0.013/0.388 & 0.500/0.500 & 1.000/1.000 & 1.000/0.975 \\
Phi-3.5-mini & 0.300/0.275 & 0.988/0.613 & 0.300/0.300 & 0.100/0.000 & 1.000/1.000 \\
\addlinespace[2pt]
\multicolumn{6}{@{}l}{Both-world validity (Flat/SVL)} \\
Qwen3-14B & 1.000/1.000 & 1.000/1.000 & 1.000/1.000 & 1.000/1.000 & 0.925/0.850 \\
Gemma-3-12B & 0.750/0.750 & 1.000/1.000 & 1.000/1.000 & 1.000/1.000 & 0.600/0.588 \\
Mistral-7B & 0.408/0.263 & 0.013/0.388 & 0.500/0.500 & 0.683/0.600 & 0.525/0.475 \\
Phi-3.5-mini & 0.300/0.275 & 0.988/0.613 & 0.150/0.150 & 0.000/0.000 & 0.500/0.500 \\
\bottomrule
\end{tabular*}

\end{table}

\section{Corruption and Repair Diagnostics}
\label{app:repair}

\subsection{Reversible state interventions}

Each corruption applies one TaskSpec-compatible edit to the compiled state.
The operator has a deterministic inverse and an edit signature used to create
a size-matched irrelevant repair.  Table~\ref{tab:app-operators} defines the
six edit types.

\begin{table}[!htbp]
\caption{\textbf{Six reversible scientific-state corruptions.}  An operator
is used only when the TaskSpec assigns it a defined scientific interpretation.}
\label{tab:app-operators}
\centering
\small
\begin{tabularx}{\textwidth}{@{}p{2.0cm}X@{}}
\toprule
Operator & State change \\
\midrule
Omit & Remove one target node and its incident relations \\
Unlink & Remove a governing relation while retaining both endpoints \\
Relabel & Replace a target type or relation label with an incompatible one \\
Merge & Collapse two distinct scientific objects into one identity \\
Duplicate & Add a second copy that creates ambiguous identity or evidence use \\
Inject & Add a plausible but irrelevant node or relation \\
\bottomrule
\end{tabularx}
\end{table}

The 24-task matrix contains 60 compatible TaskSpec--operator pairs.  Each
task--seed--reader contributes one shared intact baseline, while each
compatible pair adds three intervention worlds: target corruption, exact
target repair, and a matched edit at an unrelated location.  Thus the matrix
contains $24\times6\times4 + 60\times6\times4\times3 = 4{,}896$
decisions.  Exact repair restores the original canonical state.  The matched
condition leaves the decisive damage in place, separating recovery caused by
the governing information from recovery caused by a generic edit.

\subsection{Repair specificity and operator fingerprints}

Table~\ref{tab:app-repair-modules} contrasts exact and irrelevant repair for
each reader and module.

\begin{table}[!htbp]
\caption{\textbf{Exact target repair minus matched irrelevant repair.}
Module columns reveal where the reader-specific recovery originates.}
\label{tab:app-repair-modules}
\centering
\small
\begin{tabular}{@{}lrrrrrr@{}}
\toprule
Model & Overall & Infer. & Blind & Construct & Absence & Sunk \\
\midrule
Qwen3-14B & +0.269 & +0.000 & +0.500 & +0.681 & +0.833 & -0.333 \\
Gemma-3-12B & +0.328 & +0.000 & +0.500 & +0.667 & +0.875 & +0.083 \\
Mistral-7B & +0.150 & +0.007 & +0.438 & +0.000 & +0.667 & +0.000 \\
Phi-3.5-mini & -0.003 & -0.021 & +0.042 & +0.000 & +0.000 & +0.000 \\
\bottomrule
\end{tabular}

\end{table}

Exact repair restores 26.9 points for Qwen, 32.8 for Gemma, and 15.0 for
Mistral relative to a matched irrelevant edit.  Phi changes by -0.3 points.
Blind and Absence damage are recoverable in the first three readers, whereas
Sunk Cost and Inferential repairs expose weaker or reader-specific dependence.

For each operator, a ten-dimensional vector records target-versus-corrupted
and target-versus-matched effects across the five modules.  Missing
module--domain cells contribute zeros.  For each operator, we compute the
centroid over nonzero domain vectors and report the mean cosine to that
centroid.  All-zero vectors are omitted, and an operator with fewer than two
nonzero vectors receives 0.0.  Cross-domain cosine similarity measures whether
the resulting failure pattern is stable
(Table~\ref{tab:app-operator-stability}).

\begin{table}[!htbp]
\caption{\textbf{Cross-domain corruption-fingerprint stability.}  A star
marks cosine similarity of at least 0.50.}
\label{tab:app-operator-stability}
\centering
\small
\begin{tabular}{@{}lrrrrrr@{}}
\toprule
Model & Inject & Merge & Omit & Relabel & Unlink & Duplicate \\
\midrule
Qwen3-14B & 1.000$^{*}$ & 1.000$^{*}$ & 1.000$^{*}$ & 0.789$^{*}$ & 0.866$^{*}$ & 0.000 \\
Gemma-3-12B & 0.942$^{*}$ & 1.000$^{*}$ & 1.000$^{*}$ & 0.789$^{*}$ & 1.000$^{*}$ & 0.000 \\
Mistral-7B & 0.000 & 0.000 & 0.809$^{*}$ & 0.830$^{*}$ & 0.000 & 0.000 \\
Phi-3.5-mini & 0.000 & 0.000 & 0.628$^{*}$ & 0.000 & 0.397 & 0.333 \\
\bottomrule
\end{tabular}

\end{table}

Qwen and Gemma retain stable fingerprints for five operators, Mistral for Omit
and Relabel, and Phi for Omit.  Value pairs and repairs thus answer different
questions: the former tests whether an intact reader follows a changed state;
the latter asks whether restoring a damaged component selectively recovers
valid decisions.

\section{Source and Serialization Controls}
\label{app:source-controls}

\subsection{Upstream intervention design}

The source-control evaluation changes one decisive value in the canonical
public record before constructing any model input.  Both worlds are then
rendered as Raw Events, BM25 Events, Event Graph, Matched Source, Flat SSG,
typed SVL, Opaque Flat, and Opaque SVL.  Every view therefore inherits the
same source pair, oracle relation, seed, request, and legal choices.

Matched Source exposes the complete public source-fact closure while forbidding
compiler-derived fields.  Neutral padding equalizes its full-chat token count
with Flat SSG separately for every reader, TaskSpec, seed, and world.  Opaque
views replace state values, relation names, and action labels through a
TaskSpec-specific codebook shared across both worlds and serializations.  The
two controls isolate compiled relations from input length and familiar
scientific terminology.  Table~\ref{tab:app-source-contrasts} reports the five
contrasts needed to separate these explanations.

\begin{table}[!htbp]
\caption{\textbf{Source- and serialization-level contrasts.}  Effects are
balanced across 12 module--design families.  Brackets give 95\% family-resampling
intervals.  Matched Source and Flat SSG have equal full-chat token counts.}
\label{tab:app-source-contrasts}
\centering
\small
\setlength{\tabcolsep}{2pt}
\begin{tabular*}{\textwidth}{@{\extracolsep{\fill}}lccc@{}}
\toprule
Reader & Flat$-$Event & Flat$-$Matched Source & Opaque Flat$-$Event \\
\midrule
Qwen3-14B & +0.174 [+0.056, +0.300] & +0.194 [+0.066, +0.335] & +0.099 [$-$0.066, +0.257] \\
Gemma-3-12B & +0.133 [+0.000, +0.273] & +0.011 [$-$0.136, +0.162] & +0.280 [+0.153, +0.405] \\
Mistral-7B & +0.207 [+0.078, +0.347] & +0.193 [+0.064, +0.333] & +0.115 [+0.031, +0.215] \\
Phi-3.5-mini & +0.189 [+0.045, +0.375] & $-$0.295 [$-$0.460, $-$0.099] & +0.550 [+0.439, +0.668] \\
\bottomrule
\end{tabular*}

\begin{tabular*}{\textwidth}{@{\extracolsep{\fill}}lcc@{}}
\toprule
Reader & SVL$-$Flat & Opaque SVL$-$Opaque Flat \\
\midrule
Qwen3-14B & +0.003 [+0.000, +0.010] & $-$0.085 [$-$0.167, $-$0.019] \\
Gemma-3-12B & +0.240 [+0.125, +0.356] & $-$0.017 [$-$0.087, +0.035] \\
Mistral-7B & $-$0.005 [$-$0.023, +0.009] & $-$0.029 [$-$0.072, +0.000] \\
Phi-3.5-mini & +0.488 [+0.276, +0.696] & $-$0.024 [$-$0.049, $-$0.003] \\
\bottomrule
\end{tabular*}
\end{table}

Flat SSG remains above Event Graph for every reader after the shared upstream
intervention.  Under exact length matching, Flat also outperforms Matched Source
for Qwen and Mistral and remains similar for Gemma.  These controls separate the
benefit of materialized scientific relations from retrieval depth and context
length.

\section{Independent Replications and Broader Memory Policies}
\label{app:replications}

\subsection{Hosted-reader replication}

Two hosted readers were evaluated on two independently generated seed blocks.
The compiled-state advantage is positive in all four reader--block
combinations, ranging from 47.0 to 48.6 points
(Table~\ref{tab:app-hosted-replication}).

\begin{table}[!htbp]
\caption{\textbf{Independent hosted-reader replication.}  Brackets give 95\%
TaskSpec-resampling intervals.  Flat SSG and typed SVL differ by at most 2.3
points.}
\label{tab:app-hosted-replication}
\centering
\small
\setlength{\tabcolsep}{3pt}
\begin{tabular*}{\textwidth}{@{\extracolsep{\fill}}llccc@{}}
\toprule
Data block & Reader & Typed SVL$-$Event Graph & Flat$-$Typed SVL & BM25$-$Raw \\
\midrule
Block 1 & gpt-5.6-sol & +0.486 [+0.434, +0.534] & $-$0.011 [$-$0.022, $-$0.002] & $-$0.053 [$-$0.083, $-$0.027] \\
Block 1 & gpt-5.5 & +0.483 [+0.430, +0.533] & $-$0.009 [$-$0.025, +0.002] & $-$0.028 [$-$0.052, $-$0.008] \\
Block 2 & gpt-5.6-sol & +0.477 [+0.425, +0.523] & $-$0.023 [$-$0.052, $-$0.002] & $-$0.039 [$-$0.070, $-$0.009] \\
Block 2 & gpt-5.5 & +0.470 [+0.419, +0.519] & $-$0.013 [$-$0.028, $-$0.002] & $-$0.022 [$-$0.039, $-$0.006] \\
\bottomrule
\end{tabular*}
\end{table}

\subsection{Independent Qwen replication}

A 4,480-decision independent Qwen evaluation reproduces the
open-weight result.  Typed SVL exceeds Event Graph by 18.6 points, while Flat
SSG and typed SVL remain within one point
(Table~\ref{tab:app-qwen-replication}).

\begin{table}[!htbp]
\caption{\textbf{Independent Qwen replication.}  Brackets give 95\%
TaskSpec-resampling intervals.}
\label{tab:app-qwen-replication}
\centering
\small
\begin{tabular*}{\textwidth}{@{\extracolsep{\fill}}lccc@{}}
\toprule
Reader & Typed SVL$-$Event Graph & Flat$-$Typed SVL & BM25$-$Raw \\
\midrule
Qwen3-14B & +0.186 [+0.120, +0.253] & +0.009 [+0.000, +0.023] & +0.008 [+0.000, +0.019] \\
\bottomrule
\end{tabular*}
\end{table}

\subsection{Additional memory-policy evaluation}

An independently constructed 24-TaskSpec set spans all five modules and six
domains.  Across four open readers and ten unseen seeds, it produces 6,720
decisions.  SVL exceeds the mean of BM25, summary, selective, and provenance
memory by 28.5, 29.9, 14.4, and 11.1 points for Qwen, Gemma, Mistral, and Phi.
The TaskSpec-resampling intervals remain positive for every reader
(Table~\ref{tab:app-policy-effects}).

\begin{table}[!htbp]
\caption{\textbf{TaskSpec-balanced effect against the four conventional
memory policies.}  The effect compares SVL with the mean of BM25, summary,
selective, and provenance conditions.}
\label{tab:app-policy-effects}
\centering
\small
\begin{tabular*}{\textwidth}{@{\extracolsep{\fill}}lrrrr@{}}
\toprule
Reader & Effect & 95\% interval & Positive tasks & Leave-one-task-out range \\
\midrule
Qwen3-14B & +0.285 & [+0.228, +0.335] & 20/24 & [+0.279, +0.298] \\
Gemma-3-12B & +0.299 & [+0.238, +0.354] & 20/24 & [+0.291, +0.312] \\
Mistral-7B & +0.144 & [+0.051, +0.238] & 12/24 & [+0.129, +0.160] \\
Phi-3.5-mini & +0.111 & [+0.060, +0.167] & 10/24 & [+0.102, +0.116] \\
\bottomrule
\end{tabular*}
\end{table}

Table~\ref{tab:app-policy-modules} decomposes the same comparison by validity
module.

\begin{table}[!htbp]
\caption{\textbf{Module-level effects on the additional TaskSpec set.}
Entries use the same SVL-versus-conventional-memory contrast as
Table~\ref{tab:app-policy-effects}.}
\label{tab:app-policy-modules}
\centering
\small
\begin{tabular*}{\textwidth}{@{\extracolsep{\fill}}lrrrrr@{}}
\toprule
Reader & Inferential & Blind & Construct & Absence & Sunk Cost \\
\midrule
Qwen3-14B & +0.292 & +0.306 & +0.375 & +0.406 & +0.000 \\
Gemma-3-12B & +0.304 & +0.306 & +0.392 & +0.444 & +0.000 \\
Mistral-7B & +0.100 & $-$0.150 & +0.429 & +0.219 & +0.000 \\
Phi-3.5-mini & +0.000 & +0.300 & +0.246 & +0.000 & +0.000 \\
\bottomrule
\end{tabular*}
\end{table}

Construct Drift and Absence Algebra yield the largest extension effects.  Sunk
Cost is tied because its correct operation is to suppress irrelevant history,
not expose more of it.  Mistral's negative Blind effect alongside its large
Construct gain further demonstrates an interaction between representation,
reader capability, and validity rule.

The structural audit in Table~\ref{tab:app-projection-audit} explains which
scientific objects and governing relations remain available under each policy.

\begin{table}[!htbp]
\caption{\textbf{Structural content preserved by each memory policy.}  The
audit covers 480 TaskSpec--seed worlds.  SVL is the only view that retains all
nodes, governing relations, and complete relation endpoints.}
\label{tab:app-projection-audit}
\centering
\small
\begin{tabular*}{\textwidth}{@{\extracolsep{\fill}}lrrrrrr@{}}
\toprule
Policy & Nodes & Edges & Node recall & Edge recall & Required-edge recall & Endpoints \\
\midrule
None & 0.00 & 0.00 & 0.000 & 0.000 & 0.000 & 0.000 \\
BM25 & 2.65 & 0.68 & 0.601 & 0.246 & 0.186 & 0.000 \\
Summary & 2.25 & 0.67 & 0.510 & 0.271 & 0.111 & 0.000 \\
Selective & 1.58 & 0.42 & 0.379 & 0.167 & 0.111 & 0.000 \\
Provenance & 4.46 & 1.04 & 1.000 & 0.368 & 0.000 & 1.000 \\
ECM & 3.96 & 1.04 & 0.886 & 0.368 & 0.000 & 0.556 \\
SVL & 4.46 & 2.75 & 1.000 & 1.000 & 1.000 & 1.000 \\
\bottomrule
\end{tabular*}
\end{table}

Provenance preserves every node but none of the required governing edges.
Evidence-Carrying Memory (ECM) retains 88.6\% of nodes and scores 55.6\% on
Endpoints, the fraction of missing required edges with both endpoints visible
(1.0 when none are missing).  The behavioral results measure whether each
reader can exploit the structure that remains.
\section{Observed Boundaries and Extensions}
\label{app:boundaries}

The best serialization varies by reader even when compiled content is fixed.
Flat and typed views are nearly equivalent for Qwen and Mistral in the
source-control evaluation, whereas typed SVL substantially improves Gemma and
Phi.  Scientific-memory evaluation should therefore vary both state content
and reader-facing representation.
Single-world validity can overstate state use.  Mistral and Phi retain moderate
validity on compiled views while their counterfactual relation accuracy falls
well below Qwen and Gemma.  Paired interventions separate sensitivity to the
decisive value from an action that is merely plausible in one world.

More retained history is not always beneficial.  Sunk Cost tasks require the
reader to exclude past effort when future payoffs are unchanged.  Future
compilers should represent relevance and exclusion constraints, not only add
facts and relations.
The six domains provide controlled surfaces for shared validity rules.  Longer
histories, multimodal records, laboratory handoffs, and field-specific
simulators can broaden ecological coverage without changing the TaskSpec
oracle or intervention interface.  Learned ingestion and memory-writing
systems can be evaluated by comparing their induced state and downstream
decisions with these contracts.

\end{document}